\documentclass[conference]{IEEEtran}

\usepackage{cite}
\usepackage{amsmath,amssymb,amsfonts}
\usepackage{algorithm,algorithmic}
\usepackage{graphicx}
\usepackage{multirow}
\usepackage{booktabs}
\usepackage{textcomp}
\usepackage{xcolor} % for your \hl command
\usepackage{hyperref}
\hypersetup{hidelinks=true}
\usepackage{eso-pic}
\AddToShipoutPictureBG*{%
  \AtPageLowerLeft{%
    \put(0,12){%
      \parbox{\paperwidth}{\centering\footnotesize\textit{This work has been submitted to the IEEE for possible publication. Copyright may be transferred without notice, after which this version may no longer be accessible.}}%
    }%
  }%
}

\title{Nested Radially Monotone Polar Occupancy Estimation: Clinically-Grounded Optic Disc and Cup Segmentation for Glaucoma Screening}

\author{%
\IEEEauthorblockN{Rimsa Goperma\IEEEauthorrefmark{1},
Rojan Basnet\IEEEauthorrefmark{1}, and Liang Zhao\IEEEauthorrefmark{1}}
\IEEEauthorblockA{\IEEEauthorrefmark{1}Graduate School of Advanced Integrated Studies in Human Survivability (GSAIS),\\
Kyoto University, Kyoto 606-8306, Japan\\
Email: rimsa.goperma.53c@st.kyoto-u.ac.jp,
basnet.rojan.55i@st.kyoto-u.ac.jp,
liang@gsais.kyoto-u.ac.jp}
}

\begin{document}

\maketitle

% =========================================================================
%  ABSTRACT
% =========================================================================
\begin{abstract}
Valid segmentation of the optic disc~(OD) and optic cup~(OC) from
fundus photographs is essential for glaucoma screening.
Unfortunately, existing deep learning methods do not guarantee clinical
validness including star-convexity and nested structure of OD and OC,
resulting corruption in diagnostic metric, especially under
cross-dataset domain shift.
To adress this issue, this paper proposed \textbf{NPS-Net} (Nested Polar
Shape Network), the first framework that formulates the OD/OC segmentation
as \emph{nested radially monotone polar occupancy estimation}.
This output representation can guarantee the aforementioned clinical
validness and achieve high accuracy.
Evaluated across seven public datasets, NPS-Net
shows strong zero-shot generalization.
On RIM-ONE, it maintains 100\% anatomical validity
and improves Cup Dice by 12.8\% absolute over the best
baseline, reducing vCDR MAE by over 56\%. On PAPILA,
it achieves Disc Dice of 0.9438 and Disc HD95 of
2.78\,px, an 83\% reduction over the best competing method.
%These results demonstrate that the proposed method can
%deliver clinical reliability much more than existing methods.
\end{abstract}

\begin{IEEEkeywords}
Glaucoma Screening, optic disc and cup segmentation,
polar occupancy estimation, neuroretinal rim profiling, zero-shot
generalization, test-time adaptation.
\end{IEEEkeywords}

% =========================================================================
\section{Introduction}
\label{sec:introduction}
% =========================================================================

\IEEEPARstart{G}{laucoma} is a progressive optic neuropathy and the
leading cause of irreversible blindness, affecting over 80~million
individuals with projections exceeding 111~million by
2040~\cite{bib1,bib2}.  Because lost vision cannot be recovered, early
structural detection is paramount. For that purpose,
the \textit{vertical cup-to-disc
ratio~}(vCDR) and the \textit{angular neuroretinal rim~}(NRR) thickness profile
serve as the primary biomarkers for screening and progression
monitoring~\cite{bib3,bib4}.
Both of them require valid delineation of the
\textit{optic dis}c~(OD) and \textit{optic cup}~(OC) from color fundus
photographs~\cite{bib5}.

Many deep learning models have been proposed for OD/OC
segmentation.
Some of them can achieve high Dice
scores on benchmarks such as REFUGE and
DRISHTI-GS~\cite{bib6,bib7}.  However, they optimize
pixel-level overlap over unstructured binary masks in Cartesian
space with no guarantee on clinical validness.
Particularly, they may fail to find \emph{star-convex}
OD and OC boundaries, where ``star-convex'' means any line from
the optic nerve head crosses the boundary exactly once, a
common assumption used in practice.
Moreover, they do not guarantee that
the cup is nested in the disc.
%\emph{Third}, the clinically actionable quantity is not the mask
%itself but the angular rim profile, whose
%spatial pattern of thinning distinguishes early focal from advanced
%diffuse glaucomatous damage. Under cross-dataset domain shift, this
These weakness manifests as clinical failures such as cup-outside-disc
violations, fragmented contours, uninterpretable vCDR,
and indefinable NRR, which are not considered by Dice scores.

To mitigate these failures, some methods have been proposed including the
introduction of boundary-aware losses~\cite{bib14,bib15,bib16,bib17}, adversarial shape
constraints~\cite{bib17,bib18}, and polar coordinate
preprocessing~\cite{bib19,bib20}. These approaches try to improve
anatomical plausibility by training-time regularizers.
However, since they predict with no hard constraint,
clinical failures remain possible at the inference step,
especially under cross-dataset domain shift.
On the other hand, domain adaptation~\cite{bib21,bib22,bib23} and test-time
adaptation~\cite{bib24,bib25,bib26} methods target distributional
shift but either require target-domain data or iterative parameter
updates not tailored to failures due to scale variation
and off-center disc placement which dominate cross-dataset fundus
evaluation.

We contend that the core limitation in the literature
is the output representation itself. Instead, we
propose \textbf{NPS-Net} (Nested Polar Shape Network) that predicts
two normalised radial boundary functions $r_c(\theta)$ and $r_d(\theta)$ for the cup and the disc respectively in a
disc-centered polar domain such that
$0\!\leq\!r_c(\theta)\!\leq\!r_d(\theta)\!\leq\!1$ holds for all $\theta$.
To our knowledge,
this is the first presentation that guarantees both star-convexity and nested
containment of the output for OD/OC segmentation.  The contributions are fourfold:

\begin{enumerate}
\item \textbf{Monotone polar occupancy.}  The proposed
      cumulative-decrement construction enforces radial
      star-convexity in the output space, whereas all prior polar-coordinate
      methods~\cite{bib19,bib20} only apply the transform as a
      preprocessing step and retain unconstrained mask outputs.

\item \textbf{Factorized nesting.}  We propose a multiplicative
      gating mechanism to avoid
      cup-outside-disc output. To our knowledge,
      this is the first \emph{hard} topological
      guarantee for joint OD/OC
      segmentation, superseding existing soft-constraint
      approaches~\cite{bib14,bib15,bib16,bib17,bib18}.

\item \textbf{Entropy-gated shape prior fusion.}  We design a
      dual-path decoder that blends dense local features with a
      global angular shape prior through a confidence gate derived
      from boundary-distribution entropy, providing automatic,
      parameter-free fallback to geometric structure when local
      appearance degrades---a mechanism distinct from existing
      test-time adaptation methods that require iterative parameter
      updates~\cite{bib24,bib25}.

\item \textbf{Polar Test-Time Augmentation (Polar-TTA).}  We develop
      a training-free inference strategy that corrects scale and
      centration misalignment by evaluating a structured hypothesis
      grid, selected by internal model confidence with no parameter
      update.
\end{enumerate}

To study the proposed method, we validate it
across seven datasets spanning five
ethnicities and seven camera systems
(Table~\ref{tab:datasets}).  Under zero-shot evaluation on RIM-ONE,
NPS-Net improves Cup Dice by 12.8\% absolute over
BEAL~\cite{bib17}, reduces vCDR MAE by 56\%, and achieves 100\%
anatomical validity.  On PAPILA~\cite{bib35}, NPS-Net
attains Disc Dice of 0.9438 and Disc HD95 of 2.78\,px---over 83\%
boundary precision improvement relative to the best competing
baseline.

The rest of the paper is organized as follows.
We reivew the literature in Section~\ref{sec:related},
then present the methodology, empiricial results, and discussions
in Sections~\ref{sec:method}, \ref{sec:results} and \ref{sec:discussion},
respectively.
Finally we conclude in Section~\ref{sec:conclusion}.

% =========================================================================
\section{Literature Review}
\label{sec:related}
% =========================================================================

UNet~\cite{bib10} and its attention-augmented variants~\cite{bib11} are
popular deep learning based encoder--decoder paradigm for glaucoma screening.
Subsequent architectures introduced multi-scale feature aggregation
and channel attention~\cite{bib27,bib28}, while transformer-based
models~\cite{bib12,bib13} extended receptive fields via global
self-attention.  Fu~et~al.~\cite{bib29} demonstrated the benefits of
joint disc segmentation and glaucoma grading;
Gu~et~al.~\cite{bib30} showed that context-aware encoding sharpens
boundary delineation.  Despite strong in-domain performance, all
these methods predict OD and OC as independent classes within
unconstrained Cartesian masks, providing no structural guarantee on
nested containment.

To address this issue,
BGA-Net~\cite{bib14} and BEAC-Net~\cite{bib15} incorporate
boundary-prior branches and adversarial training for contour
fidelity.  NENet~\cite{bib16} couples a nested backbone with a patch
discriminator; BEAL~\cite{bib17} exploits entropy maps as adversarial
targets.  SCUDA~\cite{bib18} embeds a domain-invariant shape
constraint within the adaptation objective.  Critically, each operates
as a \emph{soft} training-time regularizer: the output remains an
unconstrained pixelwise mask, and nesting violations surface at
inference under domain shift.

Polar preprocessing has been explored to simplify boundary
extraction~\cite{bib19,bib20}, reformulating contour detection as 1-D
curve regression over angle.  However, these approaches retain
pixelwise mask prediction as the output target---the polar transform
serves as a preprocessing step only.  Without enforcing radial
monotonicity in the output space, predictions remain susceptible to
fragmentation.

Regarding domain shift, unsupervised adaptation methods
including feature alignment~\cite{bib21}, category-level
regularization~\cite{bib22}, transformer-based
adaptation~\cite{bib23}, and domain
disentanglement~\cite{bib31,bib32} address distributional shift at
training time.  DoFE~\cite{bib34} dynamically enriches source-domain
features with domain-specific knowledge.  Test-time approaches via
entropy minimization~\cite{bib24}, shape moment
constraints~\cite{bib25}, and atlas registration~\cite{bib26} adapt
at inference but require iterative parameter updates.  None targets
geometric canonicalization failures in polar segmentation.

% =========================================================================
\section{The Proposed Method}
\label{sec:method}
% =========================================================================

\begin{figure*}[!t]
    \centering
    \includegraphics[width=0.92\textwidth]{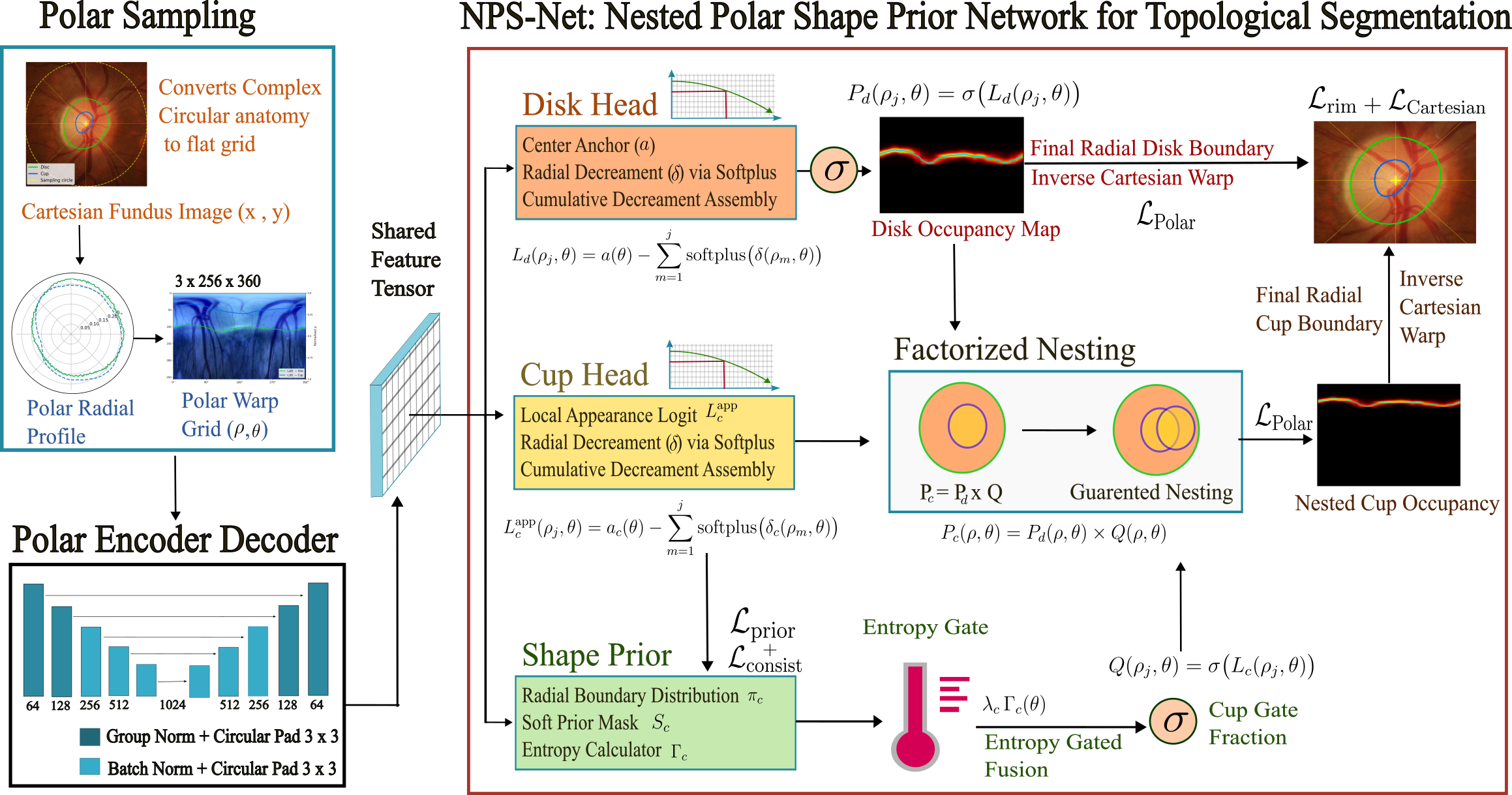}
    \caption{Overview of the proposed NPS-Net pipeline.}
    %A fundus image is
    %warped into a disc-centered polar grid via differentiable
    %bilinear sampling~\cite{bib37}.  The polar encoder--decoder extracts shared
    %features~$\mathbf{F}$ using circular padding along~$\theta$.
    %Two prediction paths operate on~$\mathbf{F}$: (i)~monotone
    %occupancy heads produce $P_d$ and $Q$ via cumulative-decrement
    %logits, with $P_c\!=\!P_d\!\times\!Q$ guaranteeing nesting;
    %(ii)~the shape prior branch predicts boundary distributions whose
    %confidence gates fusion with the dense path.  Polar-TTA evaluates
    %a hypothesis grid at inference, selecting the best alignment by
    %internal model confidence.}
    \label{fig:architecture}
\end{figure*}

Our approach does not add geometric constraints atop a mask predictor.
It changes \emph{what is predicted}.
The idea is to predict in a coordinate system that guarantees
structural constraints by construction.
The pipeline is illustrated in Fig.~\ref{fig:architecture} and
explaind in the following subsections.
%pipeline comprises five components: disc-centered polar sampling
%(Sec.~\ref{sec:polar_sampling}), a polar encoder--decoder
%(Sec.~\ref{sec:encoder}), monotone occupancy heads with factorized
%nesting (Secs.~\ref{sec:monotone}--\ref{sec:nesting}), entropy-gated
%shape prior fusion (Sec.~\ref{sec:shape_prior}), and Polar Test-Time
%Augmentation (Sec.~\ref{sec:polar_tta}).

% We first establish the geometric definitions upon which the subsequent
% subsections build.

% -------------------------------------------------------------------------
\subsection{Disc-Centered Polar Sampling}
\label{sec:polar_sampling}
% -------------------------------------------------------------------------
We assume that the input is a region-of-interest crop of $H \times W$ pixels 
centered on the optic nerve head. 
Define the polar anchor as $(c_x, c_y) = (W/2, H/2)$ and the 
normalization radius as $R = \min(H, W)/2$. 
We map a spatial point $(x, y)$ to a \emph{normalized} polar coordinate $(\rho, \theta)$ such that
\begin{equation}
    \rho = \frac{\sqrt{(x-c_x)^2 + (y-c_y)^2}}{R}, \quad
    \theta = \mathrm{atan2}(y-c_y, x-c_x).
    \label{eq:def_polar}
\end{equation}
The value $\rho \ge 0$ is called the normalized radial distance and 
$\theta \in \mathbb{S}^1 = (-\pi, \pi]$ the polar angle. Assuming that the OD and OC are star-convex 
with respect to the polar anchor,
%---meaning every ray crosses each boundary exactly once---
we introduce two \emph{boundary radius functions}
\begin{equation}
	r_d(\theta),\; r_c(\theta) : \mathbb{S}^1 \to [0,1],
    \label{eq:boundary_functions}
\end{equation}
to represent the normalised distances from the polar anchor
to the boundaries of the disc and the cup, respectively.

An \emph{occupancy map} for a boundary $r(\theta)$ is a function
$P:[0,1] \times \mathbb{S}^1 \to [0,1]$ of the probability that a polar coordinate $(\rho, \theta)$ is within the boundary $r(\theta)$.
%
%It generalizes the binary inside/outside indicator.
For a star-convex boundary with radius function $r(\theta)$, the ideal occupancy map is
$P(\rho,\theta) = \mathbf{1}[\rho \leq r(\theta)]$.
For an occupancy map $P$, we may recover the boundary $r(\theta)$ by integrating occupancy, i.e.,
\begin{equation}
    r(\theta) = \int_0^1 P(\rho,\theta)\,\mathrm{d}\rho
    \;\approx\; \frac{1}{N_\rho}\sum_{j=1}^{N_\rho} P(\rho_j,\theta),
    \label{eq:radius_from_occupancy}
\end{equation}
where $\rho_j=j/N_\rho$ for $N_\rho$ samples on $\rho$. 

We wrap the fundus crop into polar coordinates via the well-known
differentiable bilinear sampling~\cite{bib37},
producing a polar image
$\mathcal{I}_{\mathrm{pol}}\!\in\!\mathbb{R}^{3\times N_\rho\times
N_\theta}$ with $N_\rho\!=\!256$ radial bins and
$N_\theta\!=\!360$ angular bins. We use the crop center rather than ground-truth centroids
during training to ensure the model learns to tolerate standard
cropping noise.
Residual misalignment is corrected at inference by
Polar-TTA (see Subsec.~\ref{sec:polar_tta}).

A clinically valid segmentation needs to satisfy the following
anatomical invariants.

\smallskip
\noindent\textbf{(C1) Star-convexity:}  Occupancy is radially non-increasing.
\begin{equation}
    \rho' < \rho \;\Longrightarrow\; P(\rho',\theta)
    \geq P(\rho,\theta), \ \forall \theta\in\mathbb{S}^1,
    \label{eq:c1}
\end{equation}
for the boundaries of both OD and OC.

\smallskip
\noindent\textbf{(C2) Nested containment and valid rim profile:}
%The cup must lie inside the disc. This nesting guarantees that the neuroretinal rim (NRR) thickness,
\begin{equation}
    \mathrm{rim}(\theta) = r_d(\theta) - r_c(\theta) \geq 0, \ \forall\,\theta\in\mathbb{S}^1.
    \label{eq:c2}
\end{equation}
This encodes the spatial signature of glaucomatous damage (e.g., the healthy ISNT rule) and estimating this angular function faithfully is the primary clinical objective.

Existing methods try to enforce~(C1)--(C2) via soft loss penalties on
unconstrained Cartesian masks. In contrast, the proposed NPS-Net
encodes them as properties of the parameterization, making them exact
invariants for any input.

% -------------------------------------------------------------------------
\subsection{Polar Encoder--Decoder}
\label{sec:encoder}
% -------------------------------------------------------------------------

For polar encoder-decoder, a five-stage UNet processes
the polar image entirely in
$(\rho,\theta)$ space (Table~\ref{tab:architecture}).  Each stage
comprises two $3\!\times\!3$ convolutions using
\texttt{CircularPadConv2d}: the last $\lfloor k/2\rfloor$ angular
bins wrap to the beginning and vice versa, ensuring seamless feature
extraction across the $360^\circ$ boundary.  Zero padding is
maintained along~$\rho$, which represents the true origin and outer
crop boundary.  Without circular padding, a spurious discontinuity at
$\theta = \pm\pi$ would corrupt boundary predictions near the
wrap-around.

Stages~1--2 use GroupNorm (8~groups), whose per-sample statistics
remain stable under variable batch compositions during cross-dataset
inference---a deliberate choice for out-of-distribution robustness.
Stages~3--5 use BatchNorm for training efficiency.  The decoder
mirrors the encoder with transposed-convolution upsampling and skip
connections, producing shared features
$\mathbf{F}\!\in\!\mathbb{R}^{64\times N_\rho\times N_\theta}$
consumed by all prediction heads.  The full model contains $31.28$M
parameters, of which 99.2\% reside in the encoder--decoder; the
prediction heads are lightweight by design.

\begin{table}[!t]
\renewcommand{\arraystretch}{1.10}
\caption{Architecture and Parameter Budget}
\label{tab:architecture}
\centering
\scriptsize
\setlength{\tabcolsep}{3pt}
\begin{tabular}{l c c c c}
\toprule
\textbf{Stage} & \textbf{Channels} & \textbf{Norm}
& \textbf{Convolution} & \textbf{Params} \\
\midrule
Enc\,1 & $3\!\to\!64$    & GN\,(8) & $2\!\times\!$CircPad $3\!\times\!3$ & \multirow{5}{*}{31.04\,M} \\
Enc\,2 & $64\!\to\!128$  & GN\,(8) & $2\!\times\!$CircPad $3\!\times\!3$ & \\
Enc\,3 & $128\!\to\!256$ & BN      & $2\!\times\!$CircPad $3\!\times\!3$ & \\
Enc\,4 & $256\!\to\!512$ & BN      & $2\!\times\!$CircPad $3\!\times\!3$ & \\
Bottleneck & $512\!\to\!1024$ & BN & $2\!\times\!$CircPad $3\!\times\!3$ & \\
\midrule
Shape Prior  & --- & BN1d & $3\!\times\!$Circ1D & 0.24\,M \\
Monotone Heads & --- & --- & Conv1d + Conv2d & $<$1\,K \\
Fusion ($\lambda_c$) & --- & --- & --- & 1 \\
\midrule
\textbf{Total} & & & & \textbf{31.28\,M} \\
\bottomrule
\end{tabular}
\end{table}

% -------------------------------------------------------------------------
\subsection{Monotone Occupancy}
\label{sec:monotone}
% -------------------------------------------------------------------------
Since a standard sigmoid output provides no guarantee on radial
monotonicity, we enforce constraint~\eqref{eq:c1} through a
\emph{cumulative-decrement} construction.
%
%See Fig.~\ref{fig:monotone_nesting} for an illustration.

\begin{figure}[!t]
    \centering
    \includegraphics[width=0.85\columnwidth,keepaspectratio]{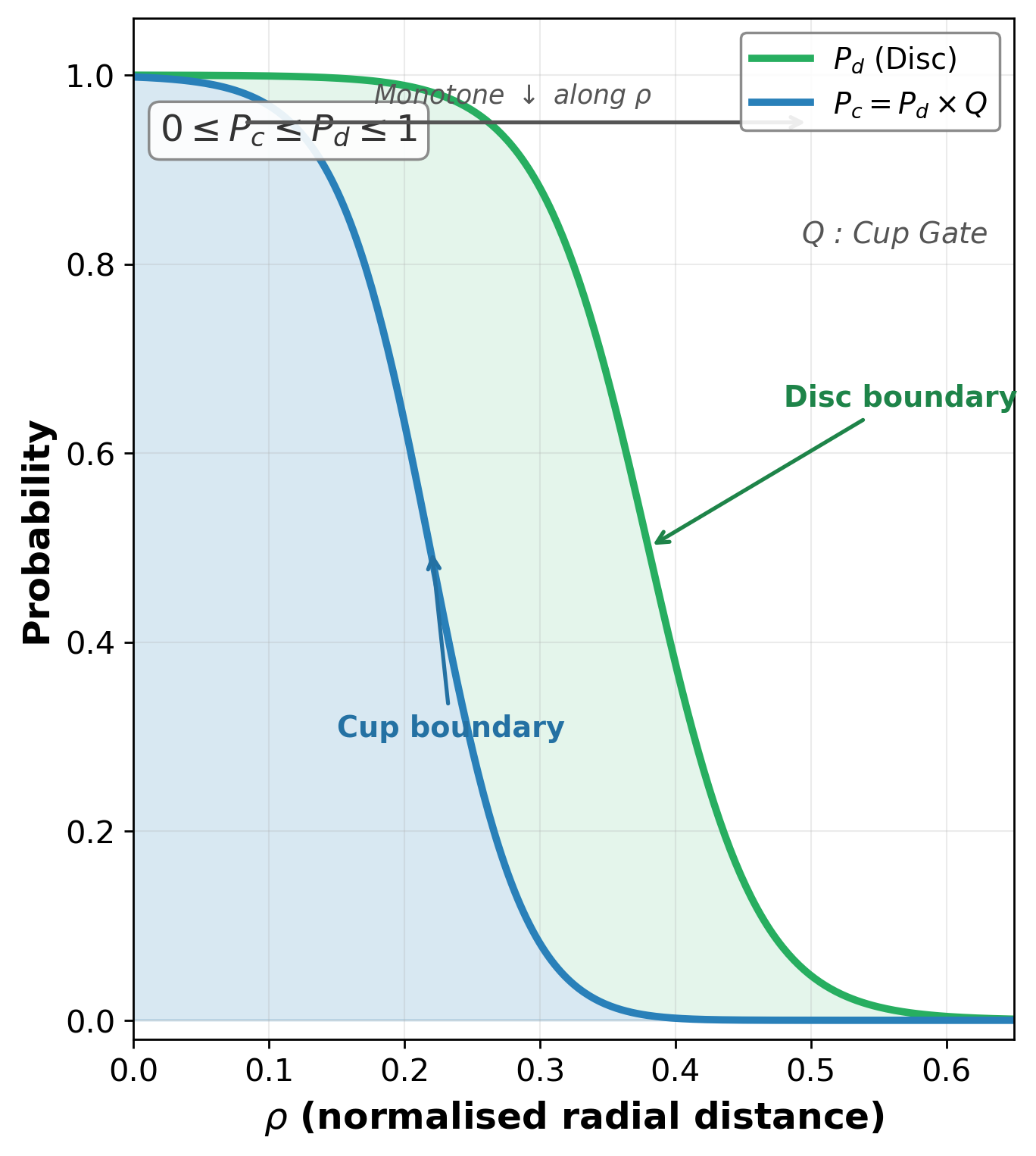}
    \caption{Illustration of monotone occupancy and factorized nesting.}
    \label{fig:monotone_nesting}
    \vspace{-1pt}
\end{figure}

Given $\mathbf{F}$, the disc head predicts two quantities:
(i)~a per-angle bias $a(\theta)\!\in\!\mathbb{R}$ via adaptive
average pooling over~$\rho$ followed by a $1\!\times\!1$ convolution,
encoding how confidently the prediction starts ``inside'' at the disc
center; and (ii)~a per-pixel non-negative decrement
$\delta(\rho,\theta)\!\geq\!0$ via a separate $1\!\times\!1$
convolution passed through
$\mathrm{softplus}(x)=\log(1+e^x)>0$, modeling how rapidly
occupancy falls toward the periphery.  The disc logit at radial
bin~$j$ is
\begin{equation}
    L_d(\rho_j,\theta) = a(\theta) -
    \sum_{m=1}^{j}
    \mathrm{softplus}\!\bigl(\delta(\rho_m,\theta)\bigr).
    \label{eq:monotone}
\end{equation}
Since $L_d(\rho_{j+1},\theta) - L_d(\rho_j, \theta) = -\mathrm{softplus}\bigl(\delta(\rho_{j+1},\theta)\bigr) < 0,$
$L_d(\rho_j,\theta)$ decreases monotonically in~$j$ for any fixed
$\theta$.  Let
\begin{equation}
    P_d(\rho_j,\theta) = \sigma\!\bigl(L_d(\rho_j,\theta)\bigr),
    \label{eq:disc_occ}
\end{equation}
where $\sigma$ denotes the sigmoid function.
The monotony of $P_d$ for fixed $\theta$ guarantees the star-convexity
constraint~(C1) (see Fig.~\ref{fig:monotone_nesting}).
Moreover, the cumulative sum
in~\eqref{eq:monotone} reduces to a single \texttt{cumsum} along
$\rho$, adding negligible computational cost.

\smallskip\noindent\textbf{Boundary extraction.}
Since $P_d$ is monotone for a fixed $\theta$, the boundary radius can be calculated as:
\begin{equation}
    r_d^{m}(\theta) = \frac{1}{N_\rho} \sum_{j=1}^{N_\rho} P_d(\rho_j, \theta),
    \label{eq:dense_radius}
\end{equation}
where the superscript $m$ denotes the monotone (dense) path. For the shape prior branch, which outputs a probability mass function $p_d(\rho|\theta)$ over radial bins, a soft-argmax yields a differentiable boundary estimate:
\begin{equation}
    r_d^{s}(\theta) = \sum_{j=1}^{N_\rho} \rho_j \cdot p_d(\rho_j|\theta),
    \label{eq:softargmax}
\end{equation}
with superscript~$s$ denoting the shape prior path.
Notice that both estimates are fully differentiable.

% -------------------------------------------------------------------------
\subsection{Factorized Nesting}
\label{sec:nesting}
% -------------------------------------------------------------------------
Independent prediction of disc and cup may allow the cup to exceed the
disc---a violation of~\eqref{eq:c2} that
corrupts vCDR and rim estimates.  We prevent this by letting
\begin{equation}
    P_c(\rho,\theta) = P_d(\rho,\theta) \times Q(\rho,\theta),
    \label{eq:nesting}
\end{equation}
where $Q\!\in\![0,1]$ is a learned cup gate produced by a second
monotone head (identical architecture, shared features~$\mathbf{F}$)
followed by sigmoid.  Since both $P_d$ and $Q$ lie in $[0,1]$, the
product satisfies $P_c\!\leq\!P_d$ everywhere for any
input, ensuring \eqref{eq:c2} (Fig.~\ref{fig:monotone_nesting}).  
Critically, because $Q$ inherits the same monotone construction as $P_d$,
it is also radially non-increasing, ensuring $P_c$ satisfies
star-convexity~(C1) and inherits the nesting guarantee.
Visually, the cup contour is guaranteed to lie strictly inside the disc
boundary, eliminating the ``cup leak'' artifacts visible in baseline
methods (see Fig.~\ref{fig:qualitative} for some examples).

The cup boundary radius is defined analogously:
\begin{equation}
    r_c^{m}(\theta) = \frac{1}{N_\rho}\sum_{j=1}^{N_\rho}
    P_c(\rho_j,\theta).
    \label{eq:cup_radius}
\end{equation}
By design, we have $r_c^{m}(\theta)\!\leq\!r_d^{m}(\theta)$ for all $\theta$.

% -------------------------------------------------------------------------
\subsection{Entropy-Gated Shape Prior Fusion}
\label{sec:shape_prior}
% -------------------------------------------------------------------------

Dense local features yield precise boundaries when appearance is
reliable but degrade under distribution shift.  A global shape prior
is more robust to appearance variation but lacks spatial precision.
Therefore, we gate their contributions by the model's own prediction confidence.

\smallskip\noindent\textbf{Shape prior branch.}
The features $\mathbf{F}$ are mean-pooled over $\rho$ to yield per-angle descriptors, which are processed by three circular 1-D convolutions (kernels 5, 5, 3; 128 channels; BatchNorm and ReLU) into two distributions: $p_d(\rho|\theta)$ for the disc boundary and $p_\alpha(\rho|\theta)$ for the angular cup-to-disc ratio, obtained via temperature-scaled softmax with $T=0.5$. 

Boundary radii are recovered via soft-argmax~\eqref{eq:softargmax}:
\begin{equation}
    r_d^{s}(\theta) = \textstyle\sum_j \rho_j \, p_d(\rho_j|\theta), \quad r_c^{s}(\theta) = \alpha^{s}(\theta) \cdot r_d^{s}(\theta),
    \label{eq:shape_radii}
\end{equation}
where $\alpha^{s}(\theta) = \sum_j \rho_j \, p_\alpha(\rho_j|\theta) \in [0,1]$ is the predicted angular cup-to-disc ratio. Since $\alpha^{s} \in [0,1]$, nesting (where the cup is contained within the disc) is preserved in the prior.

Soft prior occupancy masks $S_k$ for $k \in \{d, c\}$ are then rendered using the recovered radii $r_k^s$ and rendering temperature $\tau=0.03$:
\begin{equation}
    S_k(\rho, \theta) = \frac{1}{1 + \exp\left(-\frac{r_k^s(\theta) - \rho}{\tau}\right)}.
    \label{eq:soft_prior_masks}
\end{equation}

\smallskip\noindent\textbf{Confidence gate.}
Let $p_c(\cdot|\theta)$ denote the probability mass function (PMF) for the cup over radial bins for a fixed $\theta$.
The sharpness of $p_c(\cdot|\theta)$
is quantified by the normalised entropy
\begin{equation}
    \Gamma_c(\theta) = 1 +
    \frac{1}{\log N_\rho}
    \sum_{j} p_c(\rho_j|\theta) \log p_c(\rho_j|\theta)
    \;\in [0,1].
    \label{eq:gate}
\end{equation}
A sharp distribution (mass concentrated on few bins) yields $\Gamma_c \approx 1$ (high confidence in the shape prior); a flat distribution yields $\Gamma_c \approx 0$ (maximum uncertainty). This gates how much we trust the shape estimate: high confidence when $\rho_j$ peaks sharply, low when spread uniformly.

\smallskip\noindent\textbf{Gated fusion.}
The dense cup-gate logits $L_c^{\mathrm{app}}(\rho,\theta)$ are
modulated by the shape prior in proportion to its confidence:
\begin{equation}
    L_c(\rho,\theta) = L_c^{\mathrm{app}}(\rho,\theta)
        + \lambda_c\,\Gamma_c(\theta)\cdot
          \mathrm{logit}\!\bigl(S_c(\rho,\theta)\bigr),
    \label{eq:fusion}
\end{equation}
where $S_c(\rho,\theta)$ is defined in Subsection~\ref{sec:shape_prior}, $\lambda_c$ is a
learnable scalar (initialized to 0.1), and
$\mathrm{logit}(p)=\log(p/(1\!-\!p))$.  In-domain, $\Gamma_c$ is
high and the prior reinforces geometric structure.
Out-of-distribution, where local features degrade,
$\Gamma_c\!\to\!0$ and the fusion weight vanishes---the model falls
back entirely to the dense path, automatically and without parameter
updates.  The final cup gate is $Q=\sigma(L_c) \in [0,1]$.

% -------------------------------------------------------------------------
\subsection{Cartesian Rendering and Training Objective}
\label{sec:training}
% -------------------------------------------------------------------------

Polar occupancy maps are warped back to Cartesian space via
differentiable inverse sampling,
enabling Cartesian-space losses to backpropagate through the entire
polar network.

The training objective combines three core terms active from epoch~0:
Dice, binary cross-entropy (BCE) on both Cartesian-rendered and
polar-domain masks, and a rim-profile loss
\begin{equation}
    \mathcal{L}_{\mathrm{rim}} = 
    \mathrm{SmoothL1}\bigl(
    r_d^m(\theta) - r_c^m(\theta), \; 
    r_d^{*}(\theta) - r_c^{*}(\theta)\bigr),
    \label{eq:rim}
\end{equation}
where the superscript $(*)$ denotes the ground truth. This term directly supervises the angular NRR thickness—the primary clinical quantity of interest—instead of treating it as an implicit consequence of mask overlap.

The shape prior branch activates progressively: from epoch~20,
SmoothL1 regression on shape radii;
and a confidence-weighted dense--prior consistency term from
epoch~30.  This staged schedule ensures the dense path stabilizes
before the prior is introduced, preventing early-training
interference.  Table~\ref{tab:losses} summarizes all terms.

\begin{table}[!t]
\renewcommand{\arraystretch}{1.15}
\caption{Training Losses and Activation Schedule}
\label{tab:losses}
\centering
\small
\begin{tabular}{l l c c}
\toprule
\textbf{Loss} & \textbf{Supervises} & \textbf{$\lambda$}
& \textbf{Epoch} \\
\midrule
Dice + BCE    & Cartesian masks          & 1.0  & 0  \\
Dice + BCE    & Polar masks              & 0.7  & 0  \\
SmoothL1      & Rim profile~\eqref{eq:rim}   & 0.5  & 0  \\
Cross-entropy & Shape distributions      & 0.3  & 20 \\
SmoothL1      & Shape radii              & 0.5  & 20 \\
SmoothL1      & Contour smoothness       & 0.05 & 20 \\
SmoothL1      & Dense--prior consistency & 0.3  & 30 \\
\bottomrule
\end{tabular}
\end{table}

% -------------------------------------------------------------------------
\subsection{Polar Test-Time Augmentation}
\label{sec:polar_tta}
% -------------------------------------------------------------------------

The dominant failure mode of polar networks under cross-dataset
evaluation is geometric misalignment: variations in disc scale and
center placement shift the assumed polar origin, and every downstream
prediction inherits the error.

Polar-TTA evaluates a structured grid of 27~geometric hypotheses per
image, spanning center offsets
$(\Delta x,\Delta y)\!\in\!\{0,\pm8,\pm16\}$\,px and scale factors
$s\!\in\!\{0.85,1.0,1.15\}$ applied to~$R$.  Each hypothesis
generates a modified polar grid and a full
forward pass through the frozen model.  The best hypothesis is
selected by
\begin{equation}
    \mathcal{S}_i =
    0.4\,\overline{P}_d^{(i)}
    + 0.4\,\overline{\Gamma}_d^{(i)}
    + 0.2\,\mathcal{C}^{(i)},
    \label{eq:tta}
\end{equation}
where $\overline{P}_d^{(i)}$ is mean disc occupancy (higher when the
disc is well-centered), $\overline{\Gamma}_d^{(i)}$ is mean shape
prior confidence from~\eqref{eq:gate} (higher when the boundary
distribution is sharp), and $\mathcal{C}^{(i)}$ is disc compactness.
The top three hypotheses are blended via softmax-weighted averaging.
No gradient is computed and no parameter is modified.

% =========================================================================
\section{Experiments and Results}
\label{sec:results}
% =========================================================================

\subsection{Datasets and Evaluation Protocol}
\label{sec:datasets}

Table~\ref{tab:datasets} summarizes all datasets used in this study,
organized by their role in training and evaluation.  The experimental
design deliberately maximizes demographic and hardware diversity:
training data span Arab, Caucasian, Malay, and Nepalese populations
captured with four distinct camera systems, while zero-shot
benchmarks introduce Spanish/Hispanic cohorts and additional hardware
never seen during training.

\begin{table}[!t]
\renewcommand{\arraystretch}{1.15}
	\caption{Summary of the datasets}
%No image from the zero-shot benchmarks is
%seen during training, validation, or hyperparameter selection.}
\label{tab:datasets}
\centering
\scriptsize
\setlength{\tabcolsep}{2.5pt}
\begin{tabular}{l l l l r}
\toprule
\textbf{Dataset} & \textbf{Ethnicity} & \textbf{Hardware} & \textbf{Country} & \textbf{Images} \\
\midrule
\multicolumn{5}{l}{\textit{Internal Train / Validation / Test (70/15/15\%)}} \\
\midrule
RIGA--BinRushed & Arab       & Topcon TRC-NW7SF & Saudi Arabia & 233 \\
RIGA--Messidor  & Caucasian  & Topcon TRC-NW6   & France       & 463 \\
ORIGA           & Malay      & Canon CR-DGi     & Singapore    & 654 \\
NETRA~\cite{bib36} & Nepalese & Canon          & Nepal        & 480 \\
\midrule
\multicolumn{5}{l}{\textit{Cross-Domain Benchmark (DRISHTI-GS\,$_\text{test}$ + REFUGE Combined)}} \\
\midrule
DRISHTI-GS\,$_\text{test}$ & Indian & ---  & India        & 51 \\
REFUGE          & Asian\,(CN/MY)  & Zeiss Visucam\,500 & China / SG & 400 \\
\midrule
\multicolumn{5}{l}{\textit{Zero-Shot Generalization (Fully Held Out)}} \\
\midrule
RIM-ONE         & Spanish\,(Caucasian) & Nidek AFC-210 & Spain & 64 \\
PAPILA          & Hispanic / Spanish   & Topcon TRC-NW400 & Spain & 488 \\
\bottomrule
\end{tabular}
\end{table}

\smallskip\noindent\textbf{Internal split.}
The 1,830 images from RIGA--BinRushed, RIGA--Messidor, ORIGA, and
NETRA~\cite{bib36} are split into training~(1,281),
validation~(274), and test~(275) sets in the ratio 70:15:15, stratified by glaucoma/healthy
label.
%NETRA provides fundus images from a Nepalese population
%acquired during community-based glaucoma screening
%camps~\cite{bib36}, contributing demographic diversity absent from
%Western-centric benchmarks.

\smallskip\noindent\textbf{Cross-domain benchmark.}
The official test partition of DRISHTI-GS~\cite{bib7} (51~images) and
the full REFUGE~\cite{bib6} dataset (400~images) are combined into a
single evaluation set of 451~images to assess cross-domain performance
on distributions that partially overlap with training sources.

\smallskip\noindent\textbf{Zero-shot benchmarks.}
\textbf{RIM-ONE}~\cite{bib8,bib9} (64~images; Spanish stereo disc
photographs with markedly different acquisition hardware) and
\textbf{PAPILA}~\cite{bib35} (488~images; a multi-center Hispanic
fundus dataset spanning six clinical sites with heterogeneous camera
hardware) serve as strict zero-shot benchmarks.  No image from either
dataset is seen during training, validation, or hyperparameter
selection.

\smallskip\noindent\textbf{Metrics.}
We use $\uparrow$ ($\downarrow$) to show the higher (lower) the better.
\emph{Overlap}: Cup/Disc Dice ($\uparrow$), Cup/Disc HD95 ($\downarrow$;
Hausdorff distance at 95th percentile, pixels);
\emph{Clinical geometry}: vCDR MAE ($\downarrow$), angular Rim MAE
($\downarrow$), angular Rim Correlation ($\uparrow$; Pearson
correlation between predicted and ground-truth angular rim
profiles---the metric testing if the spatial
pattern distinguishing glaucomatous from healthy discs is preserved).

\smallskip\noindent\textbf{Baselines (seven).}
UNet~\cite{bib10}, Attention-UNet~\cite{bib11},
ResUNet, Polar~UNet (polar input, no monotone constraints),
TransUNet~\cite{bib13}, BEAL~\cite{bib17}, and DoFE~\cite{bib34}.
All share the same data split, preprocessing, augmentation, and
optimization.

% -------------------------------------------------------------------------
\subsection{Implementation Details}
\label{sec:implementation}
% -------------------------------------------------------------------------

All models were trained with AdamW (peak LR $3\!\times\!10^{-4}$, weight
decay $10^{-4}$), OneCycleLR schedule, 80~epochs, batch size~4.
Gradient clipping at max norm~1.0; mixed precision~(AMP) throughout.
Augmentation: flips, shift-scale-rotate, brightness-contrast,
Gaussian blur/noise (Albumentations).  CLAHE preprocessing; input
size $512\!\times\!512$.  Training: ${\sim}$23~min on a single NVIDIA
RTX~5090.  NPS-Net without Polar-TTA runs at 128~FPS (7.8\,ms/image),
comparable to TransUNet (125.7~FPS).  With Polar-TTA (27 forward
passes), throughput drops to 0.1~FPS, suited to offline batch
screening.

% -------------------------------------------------------------------------
\subsection{Internal Test Set}
\label{sec:internal}
% -------------------------------------------------------------------------

Table~\ref{tab:internal} reports results on the internal benchmark
(275~samples). Best scores are shown in {\bf bold}.  All methods approach saturation: Cup Dice exceeds
0.88 and Disc Dice exceeds 0.96 across the board.  DoFE achieves the
highest Cup Dice (0.8957) and the best in-domain clinical metrics
(vCDR MAE 0.0462, Rim Corr 0.6966).  NPS-Net remains competitive.
These narrow in-domain margins are not surprising.
The question is whether they survive domain shift.

\begin{table}[!t]
\renewcommand{\arraystretch}{1.12}
\caption{Internal Test Set (275 Samples).}
\label{tab:internal}
\centering
\scriptsize
\setlength{\tabcolsep}{1.8pt}
\begin{tabular}{l cc cc cc c}
\toprule
& \multicolumn{2}{c}{\textbf{Dice}$\uparrow$}
& \multicolumn{2}{c}{\textbf{HD95}$\downarrow$}
& \textbf{vCDR}$\downarrow$
& \textbf{Rim}$_\text{M}$$\downarrow$
& \textbf{Rim}$_\text{C}$$\uparrow$ \\
\cmidrule(lr){2-3}\cmidrule(lr){4-5}
& Cup & Disc & Cup & Disc & & & \\
\midrule
UNet       & .8929 & .9643 & 4.65 & 0.90 & .0480 & .0258 & .6309 \\
Attn-UNet  & .8946 & .9649 & 4.19 & 0.71 & .0481 & .0254 & .6450 \\
ResUNet    & .8941 & .9650 & 4.70 & 0.86 & .0520 & .0253 & .6408 \\
Polar UNet & .8830 & .9719 & 5.08 & 0.76 & .0527 & .0268 & .6520 \\
TransUNet  & .8852 & .9710 & 4.69 & \textbf{0.28} & .0534 & .0255 & .6194 \\
BEAL       & .8948 & .9756 & \textbf{4.00} & 0.12 & .0469 & .0229 & .6854 \\
DoFE       & \textbf{.8957} & .9747 & 4.03 & 0.20 & \textbf{.0462} & \textbf{.0226} & \textbf{.6966} \\
NPS-Net    & .8933 & \textbf{.9760} & 4.40 & 0.46 & .0490 & .0239 & .6746 \\
\bottomrule
\end{tabular}
\end{table}

% -------------------------------------------------------------------------
\subsection{Zero-Shot Generalization: RIM-ONE}
\label{sec:rimone}
% -------------------------------------------------------------------------

RIM-ONE constitutes the most challenging evaluation: fully unseen
hardware, protocol, and patient cohort.  Table~\ref{tab:rimone}
exposes a dramatic performance separation.

All Cartesian baselines collapse.  Cup Dice falls below 0.68 for
every standard method; Disc HD95 exceeds 13\,px even for TransUNet.
BEAL achieves Cup Dice 0.6782 but with $\pm$0.247 standard
deviation, indicating that many individual predictions are
near-complete failures.  DoFE degrades more severely (Cup Dice
0.5873), suggesting that its domain-knowledge pooling overfits to
source-domain statistics.

NPS-Net achieves Cup Dice 0.8064---12.8\% absolute over BEAL---with
substantially tighter distribution ($\pm$0.154), indicating not
merely higher average performance but greater prediction stability.
Disc Dice reaches 0.9140 ($\pm$0.026).

The clinical metrics amplify this gap.  NPS-Net reduces vCDR MAE from
BEAL's 0.1563 to 0.0684---a 56\% reduction (Fig. \ref{fig:vcdr}).  Rim Corr rises from
0.2688 to 0.5553: the angular rim-thinning pattern that distinguishes
glaucomatous from healthy discs is preserved with moderate fidelity
by NPS-Net but destroyed by all competing methods.  A vCDR MAE of
0.16 can shift a patient across diagnostic thresholds; 0.07 falls
within acceptable inter-observer variability.
Table~\ref{tab:significance} confirms statistical significance via
Wilcoxon signed-rank tests ($p\!<\!0.01$ on all metrics against
every baseline).

\begin{table}[!t]
\renewcommand{\arraystretch}{1.12}
\caption{RIM-ONE Zero-Shot (64 Samples). All baselines collapse under
domain shift; NPS-Net maintains robust performance.}
\label{tab:rimone}
\centering
\scriptsize
\setlength{\tabcolsep}{1.8pt}
\begin{tabular}{l cc cc cc c}
\toprule
& \multicolumn{2}{c}{\textbf{Dice}$\uparrow$}
& \multicolumn{2}{c}{\textbf{HD95}$\downarrow$}
& \textbf{vCDR}$\downarrow$
& \textbf{Rim}$_\text{M}$$\downarrow$
& \textbf{Rim}$_\text{C}$$\uparrow$ \\
\cmidrule(lr){2-3}\cmidrule(lr){4-5}
& Cup & Disc & Cup & Disc & & & \\
\midrule
UNet       & .5162 & .5733 & 19.01 & 33.00 & .1781 & .1046 & .0828 \\
Attn-UNet  & .5035 & .5399 & 20.37 & 31.26 & .1913 & .1114 & .0380 \\
ResUNet    & .4625 & .5230 & 21.33 & 38.88 & .2171 & .1114 & .1338 \\
Polar UNet & .6094 & .7285 & 19.30 & 23.50 & .1546 & .0783 & .1999 \\
TransUNet  & .6789 & .8539 & 18.53 & 13.51 & .1189 & .0646 & .2076 \\
BEAL       & .6782 & .8716 & 18.65 & 7.53  & .1563 & .0716 & .2688 \\
DoFE       & .5873 & .7977 & 20.28 & 11.63 & .1931 & .0732 & .2375 \\
NPS-Net    & \textbf{.8064} & \textbf{.9140} & \textbf{10.25} & \textbf{4.33} & \textbf{.0684} & \textbf{.0351} & \textbf{.5553} \\
\bottomrule
\end{tabular}
\end{table}

\begin{table}[!t]
\renewcommand{\arraystretch}{1.12}
\caption{Statistical Significance on RIM-ONE (Wilcoxon Signed-Rank,
$N\!=\!64$). $^{***}\!p\!<\!0.001$; $^{**}\!p\!<\!0.01$.}
\label{tab:significance}
\centering
\scriptsize
\setlength{\tabcolsep}{3pt}
\begin{tabular}{l cc cc cc}
\toprule
& \multicolumn{2}{c}{\textbf{Cup Dice}$\uparrow$}
& \multicolumn{2}{c}{\textbf{Disc Dice}$\uparrow$}
& \multicolumn{2}{c}{\textbf{vCDR MAE}$\downarrow$} \\
\cmidrule(lr){2-3}\cmidrule(lr){4-5}\cmidrule(lr){6-7}
& $\Delta$ & $p$
& $\Delta$ & $p$
& $\Delta$ & $p$ \\
\midrule
UNet      & +.29 & $^{***}$ & +.34 & $^{***}$ & $-$.11 & $^{**}$ \\
Attn-UNet & +.30 & $^{***}$ & +.37 & $^{***}$ & $-$.12 & $^{***}$ \\
ResUNet   & +.34 & $^{***}$ & +.39 & $^{***}$ & $-$.15 & $^{***}$ \\
Polar UNet& +.20 & $^{***}$ & +.19 & $^{***}$ & $-$.09 & $^{***}$ \\
TransUNet & +.13 & $^{**}$  & +.06 & $^{***}$ & $-$.05 & $^{***}$ \\
BEAL      & +.13 & $^{***}$ & +.04 & $^{***}$ & $-$.09 & $^{***}$ \\
DoFE      & +.22 & $^{***}$ & +.12 & $^{***}$ & $-$.12 & $^{***}$ \\
\bottomrule
\end{tabular}
\end{table}

% -------------------------------------------------------------------------
\subsection{Zero-Shot Generalization: PAPILA}
\label{sec:papila}
% -------------------------------------------------------------------------

PAPILA (488~images) constitutes the second zero-shot benchmark, with
greater variability in disc morphology across six clinical sites.
Table~\ref{tab:papila} reveals a pattern of baseline collapse
qualitatively similar to RIM-ONE.

Cartesian baselines achieve Cup Dice below 0.42 and Disc Dice below
0.43, with Disc HD95 exceeding 54\,px---catastrophic failure.
TransUNet leads baselines on Cup Dice but its Disc HD95 of
18.15\,px reveals substantial boundary imprecision.  BEAL and DoFE
show modest improvement (Cup Dice ${\sim}$0.52) but vCDR MAE exceeds
0.27 for all Cartesian methods, and Rim Corr falls below 0.32
universally.

\begin{figure}[!t]
    \centering
    \includegraphics[width=\columnwidth]{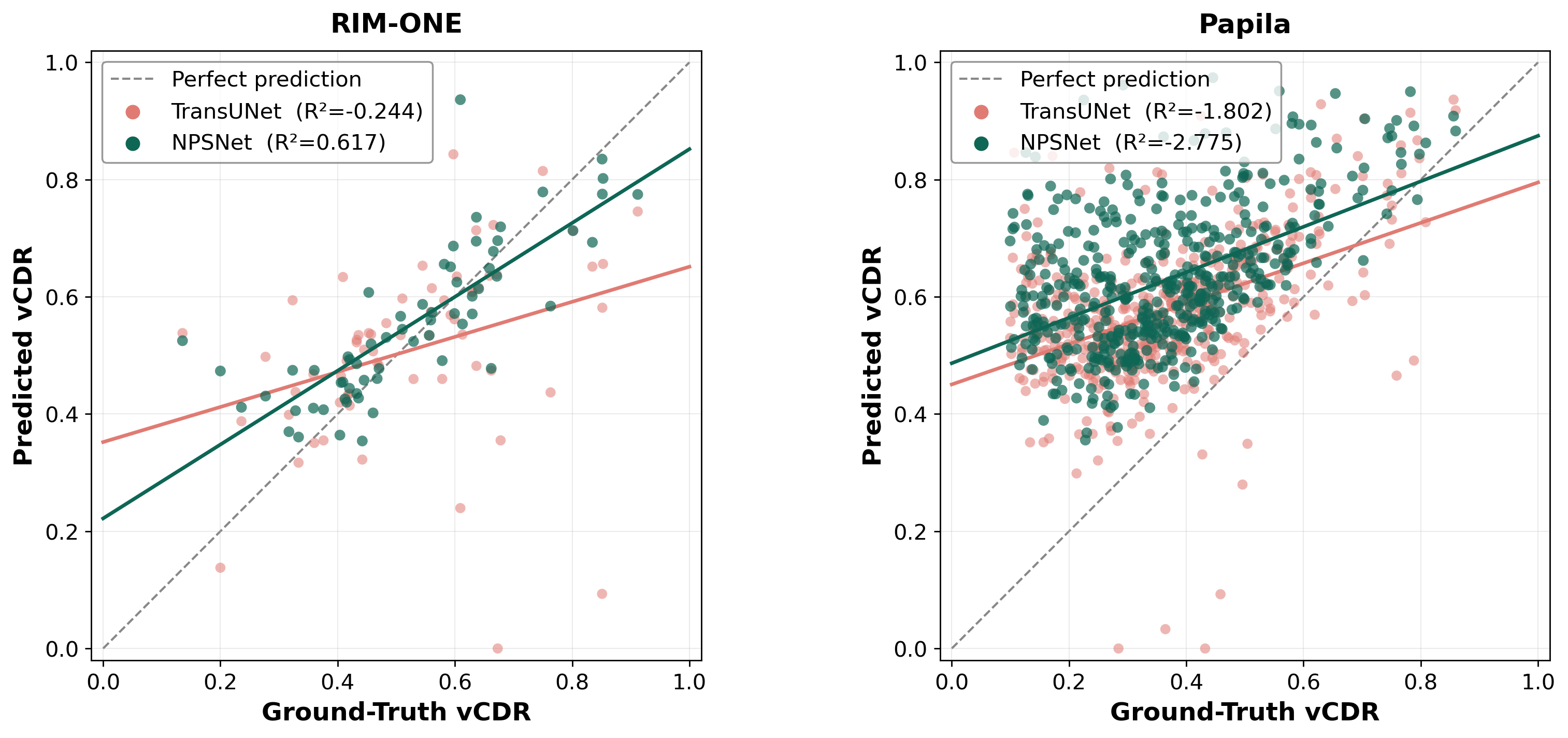}
    \caption{Predicted vs.\ ground-truth vCDR on RIM-ONE and Papila for
    TransUNet and NPS-Net.  NPS-Net clusters tightly
    along the identity line, reflecting the 56\% reduction in vCDR
    MAE on RIM-ONE and 24\% reduction on Papila dataset.}
    \label{fig:vcdr}
\end{figure}

NPS-Net achieves Disc Dice 0.9438 ($\pm$0.051)---12.5 percentage
points above the second best---and reduces Disc HD95 to
2.78\,px, over 83\% improvement relative to DoFE (16.81\,px).  Cup
Dice of 0.6063 is competitive with TransUNet, reflecting the inherent
difficulty of cup delineation in morphologically diverse discs; the
anatomical validity guarantee ensures every cup prediction remains
strictly inside its disc.  On clinical metrics, NPS-Net achieves vCDR
MAE 0.1722 (24\% reduction over TransUNet) and Rim Corr 0.5040
(64\% above TransUNet's 0.3070)  (Fig. \ref{fig:vcdr}).

The connection from methodology to metrics is direct: the monotone
construction~\eqref{eq:monotone} eliminates fragmented disc
predictions (the dominant failure), yielding the large HD95
reduction; factorized nesting~\eqref{eq:nesting} ensures valid rim
profiles even when cup appearance is ambiguous; and the entropy
gate~\eqref{eq:gate} correctly increases prior weighting under high
appearance uncertainty.

\begin{table}[!t]
\renewcommand{\arraystretch}{1.12}
\caption{PAPILA Zero-Shot (488 Samples). NPS-Net leads on disc
metrics by wide margins and achieves the best clinical geometry
scores.}
\label{tab:papila}
\centering
\scriptsize
\setlength{\tabcolsep}{1.8pt}
\begin{tabular}{l cc cc cc c}
\toprule
& \multicolumn{2}{c}{\textbf{Dice}$\uparrow$}
& \multicolumn{2}{c}{\textbf{HD95}$\downarrow$}
& \textbf{vCDR}$\downarrow$
& \textbf{Rim}$_\text{M}$$\downarrow$
& \textbf{Rim}$_\text{C}$$\uparrow$ \\
\cmidrule(lr){2-3}\cmidrule(lr){4-5}
& Cup & Disc & Cup & Disc & & & \\
\midrule
UNet       & .3786 & .4199 & 28.84 & 55.86 & .2404 & .2322 & $-$.043 \\
Attn-UNet  & .4023 & .4232 & 27.01 & 54.88 & .2694 & .2349 & $-$.074 \\
ResUNet    & .3222 & .3640 & 29.40 & 62.79 & .2689 & .2347 & .030 \\
Polar UNet & .4937 & .7316 & 33.69 & 24.93 & .2924 & .1847 & .085 \\
TransUNet  & \textbf{.6153} & .8185 & 26.71 & 18.15 & .2260 & .1441 & .307 \\
BEAL       & .5162 & .7669 & 33.63 & 23.51 & .3393 & .2018 & .163 \\
DoFE       & .5158 & .8278 & 34.09 & 16.81 & .2945 & .1755 & .250 \\
NPS-Net    & .6063 & \textbf{.9438} & \textbf{25.33} & \textbf{2.78} & \textbf{.1722} & \textbf{.1059} & \textbf{.504} \\
\bottomrule
\end{tabular}
\end{table}

% -------------------------------------------------------------------------
\subsection{Cross-Domain Benchmark: DRISHTI-GS + REFUGE}
\label{sec:drishti_refuge}
% -------------------------------------------------------------------------

Table~\ref{tab:drishti_refuge} reports all metrics on the combined
DRISHTI-GS$_\text{test}$ + REFUGE benchmark (451~images).  This
evaluation occupies an intermediate position between the internal test
set and the fully unseen zero-shot benchmarks: REFUGE shares partial
demographic overlap with ORIGA (both contain Malay subjects), yet the
combined set introduces DRISHTI-GS's Indian cohort and Zeiss Visucam
hardware not represented in training.

\begin{table}[!t]
\renewcommand{\arraystretch}{1.12}
\caption{DRISHTI-GS$_\text{test}$ + REFUGE Combined (451~Images).}
\label{tab:drishti_refuge}
\centering
\scriptsize
\setlength{\tabcolsep}{1.8pt}
\begin{tabular}{l cc cc cc c}
\toprule
& \multicolumn{2}{c}{\textbf{Dice}$\uparrow$}
& \multicolumn{2}{c}{\textbf{HD95}$\downarrow$}
& \textbf{vCDR}$\downarrow$
& \textbf{Rim}$_\text{M}$$\downarrow$
& \textbf{Rim}$_\text{C}$$\uparrow$ \\
\cmidrule(lr){2-3}\cmidrule(lr){4-5}
& Cup & Disc & Cup & Disc & & & \\
\midrule
UNet       & \textbf{.8603} & .9065 & 7.65 & 6.30  & .0858 & .0642 & .6130 \\
Attn-UNet  & .8512          & .8997 & 8.21 & 6.78  & .0744 & .0579 & .6108 \\
ResUNet    & .8497          & .9132 & 8.73 & 5.08& .0692 & .0535 & .6372 \\
Polar UNet & .8166          & \textbf{.9349} & 11.66 & \textbf{2.69} & .0859 & .0625 & .6595 \\
TransUNet  & .8565          & .8536 & 8.14 & 11.82 & .1158 & .0780 & .5717 \\
BEAL       & .8530          & .8154 & 8.06 & 16.63 & .1440 & .0951 & .4794 \\
DoFE       & .8461          & .8583 & 9.30 & 12.03 & .1344 & .0820 & .5222 \\
NPS-Net    & .8563          & .9077 & \textbf{7.54} & 6.08  & \textbf{.0636} & \textbf{.0531} & \textbf{.6856} \\
\bottomrule
\end{tabular}
\end{table}

On segmentation overlap, UNet achieves the highest Cup Dice (0.8603),
while Polar~UNet leads on Disc Dice (0.9349) and Disc HD95
(2.69\,px)---demonstrating the inherent advantage of polar-domain
processing for disc boundary localization even without monotone
constraints.  NPS-Net attains Cup Dice of 0.8563 and Cup HD95 of
7.54\,px (best overall), confirming that the monotone construction
eliminates the fragmented cup predictions that inflate boundary
distances.

Critically, the clinical geometry metrics reveal a ranking divergent
from overlap scores.  NPS-Net achieves the best vCDR MAE
(\textbf{0.0636}), the best Rim MAE (\textbf{0.0531}), and the
highest Rim Correlation (\textbf{0.6856})---sweeping all three
clinical metrics.  This dissociation between overlap and angular
fidelity is consequential: a method may achieve competitive Dice
through globally accurate masks yet misrepresent the \emph{spatial
distribution} of rim thickness.  The direct rim
supervision~\eqref{eq:rim} and factorized nesting~\eqref{eq:nesting}
give NPS-Net a decisive advantage precisely on the metrics of greatest
diagnostic relevance.

BEAL and DoFE, which led on the internal test set
(Table~\ref{tab:internal}), degrade substantially: both produce
vCDR MAE exceeding 0.13 and Rim Corr below 0.53, indicating that
their domain-adaptation mechanisms do not transfer reliably even under
moderate distributional shift.  TransUNet exhibits similar fragility
(Disc HD95 of 11.82\,px), confirming that global self-attention
without geometric structure provides limited protection against
domain mismatch.

% -------------------------------------------------------------------------
\subsection{Ablation Study}
\label{sec:ablation}
% -------------------------------------------------------------------------

Table~\ref{tab:ablation} isolates each component's contribution on
RIM-ONE (zero-shot).  All variants share the same backbone, data, and
hyperparameters; components are added incrementally.

\textbf{Monotone occupancy (M2)} produces the largest single gain:
Cup Dice rises from 0.609 to 0.760 and Cup HD95 drops from 19.30
to 13.64.  Enforcing star-convexity via~\eqref{eq:monotone}
eliminates fragmented multi-island predictions---the primary failure
mode under domain shift.

\textbf{Factorized nesting (M3)} further improves all metrics and
reduces the nesting violation rate to exactly zero.  From M3 onward,
$P_c\!\leq\!P_d$ holds for all predictions across
all 64 images---a guarantee no soft-constraint method provides.

\textbf{Entropy-gated shape prior (M4)} delivers the largest
improvement in Rim Corr (0.431$\to$0.519), the metric most
sensitive to angular boundary fidelity.  The prior regularizes the
angular profile without overriding the dense path, with the
confidence gate disambiguating local appearance in uncertain regions.

\textbf{Polar-TTA (M5)} contributes a complementary increment,
confirming that training-free geometric
canonicalization~\eqref{eq:tta} addresses a distinct failure
mode---misaligned polar grids---independent of representation
constraints.

\begin{table}[!t]
\renewcommand{\arraystretch}{1.12}
\caption{Ablation on RIM-ONE (64 Samples, Zero-Shot). Each row adds
one component.}
\label{tab:ablation}
\centering
\scriptsize
\setlength{\tabcolsep}{2.2pt}
\begin{tabular}{l c c c c c c}
\toprule
\textbf{Model}
& \textbf{Cup}$_\text{D}$$\uparrow$
& \textbf{Disc}$_\text{D}$$\uparrow$
& \textbf{HD95}$_\text{C}$$\downarrow$
& \textbf{vCDR}$\downarrow$
& \textbf{Rim}$_\text{M}$$\downarrow$
& \textbf{Rim}$_\text{C}$$\uparrow$ \\
\midrule
M1: Polar UNet      & .609 & .729 & 19.30 & .155 & .078 & .200 \\
M2: +Monotone       & .760 & .929 & 13.64 & .084 & .045 & .425 \\
M3: +Nesting$^\dag$ & .781 & .935 & 11.48 & .079 & .040 & .431 \\
M4: +Shape prior    & .783 & \textbf{.930}& 11.29 & .071 & .037 & .519 \\
M5: +Polar-TTA      & \textbf{.806} & .914& \textbf{10.25} & \textbf{.068} & \textbf{.035} & \textbf{.555} \\
\bottomrule
\multicolumn{7}{l}{\scriptsize$^\dag$Violation rate = 0\% from M3
onward ($P_c\!=\!P_d\!\cdot\!Q \Rightarrow P_c\!\leq\!P_d$
everywhere).}
\end{tabular}
\end{table}

% -------------------------------------------------------------------------
\subsection{Qualitative Analysis}
\label{sec:qualitative}
% -------------------------------------------------------------------------

\begin{figure*}[!t]
    \centering
    \includegraphics[width=0.92\textwidth]{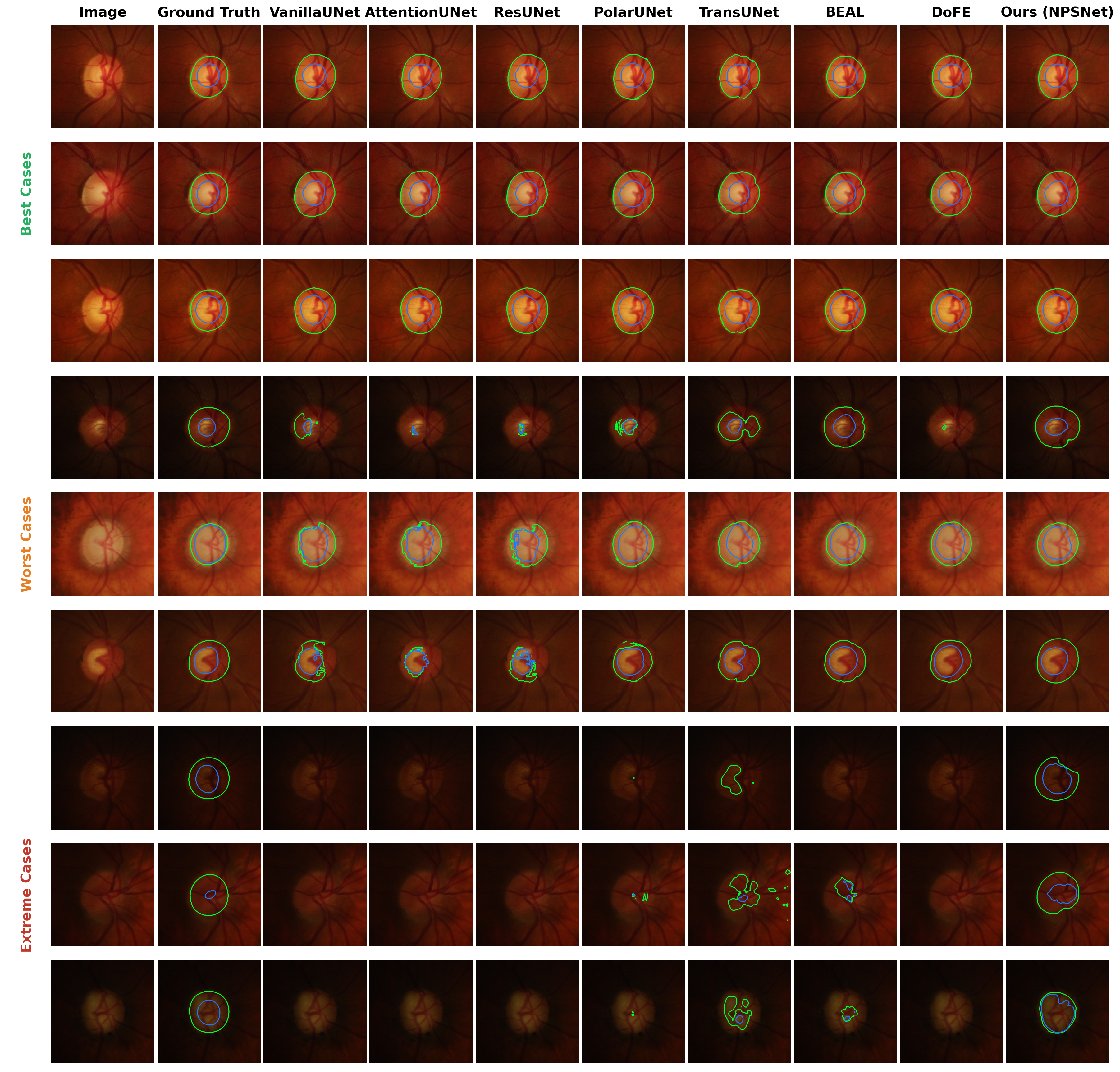}
    \caption{Qualitative comparison on RIM-ONE (zero-shot) including best
    cases (top), worst cases (middle), and extreme cases (bottom): Disc in
    green, cup in blue. }
    \label{fig:qualitative}
\end{figure*}

\begin{figure}[!t]
    \centering
    \includegraphics[width=\columnwidth]{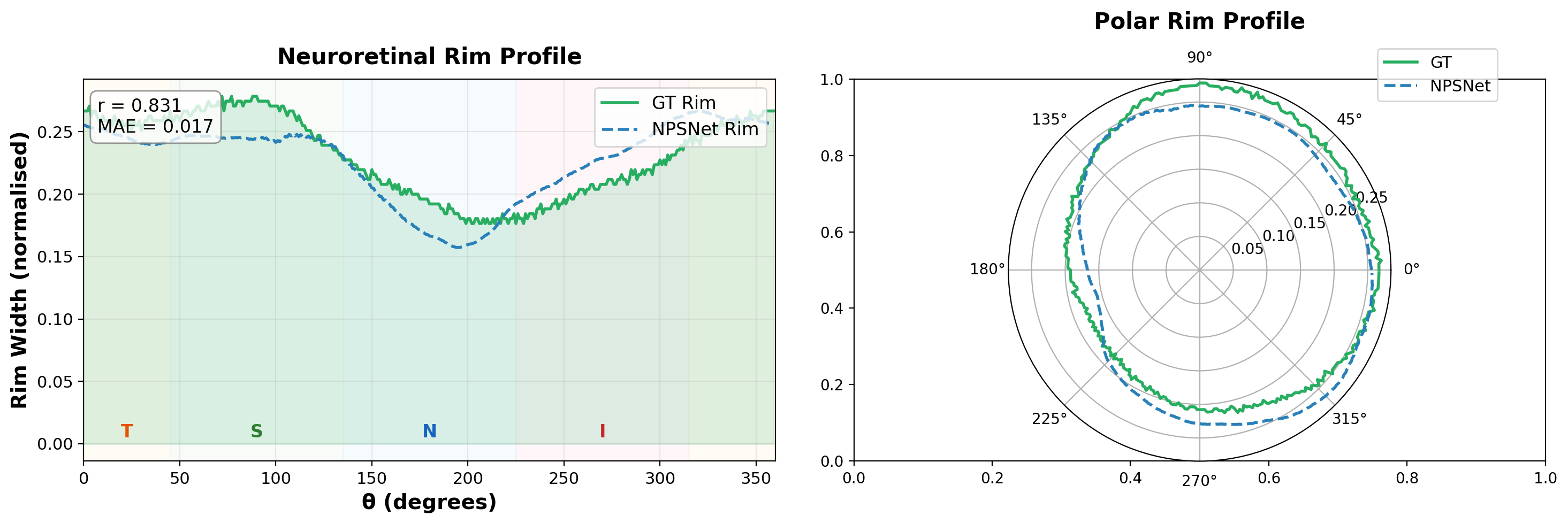}
    \caption{Angular rim profile
    $\mathrm{rim}(\theta)\!=\!r_d(\theta)\!-\!r_c(\theta)$ for a
    representative RIM-ONE image.  NPS-Net (dottet) tracks the
    ground truth (solid) with high fidelity, preserving the ISNT
    pattern.  Direct rim supervision~\eqref{eq:rim} ensures this
    is a first-class training objective.}
    \label{fig:rim}
\end{figure}

Fig.~\ref{fig:qualitative} presents segmentation overlays on RIM-ONE
across three difficulty tiers.
Under \textbf{best-case} conditions, all methods produce plausible
disc boundaries; the critical distinction lies in the cup, where
baselines generate irregular contours while NPS-Net produces
smoothly nested boundaries.  Fig.~\ref{fig:rim} confirms this
angularly: NPS-Net's predicted rim profile tracks the ground truth
with high fidelity, preserving the superior-inferior asymmetry that
is diagnostically informative.

\begin{figure}[!t]
    \centering
    \includegraphics[width=\columnwidth]{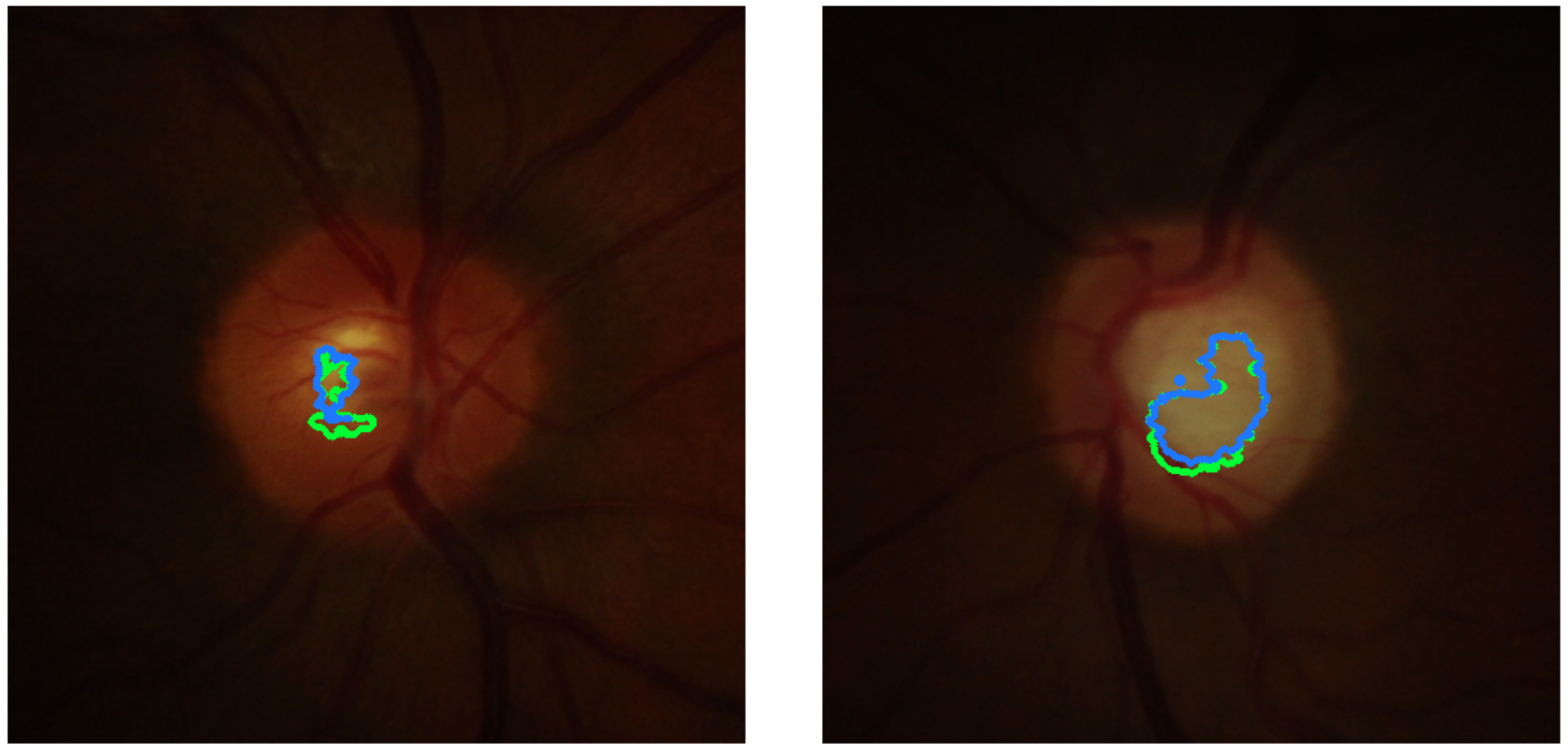}
    \caption{Representative ResUnet predictions illustrating nesting violations for the cup (blue) and the disc (green).}
    \label{fig:special}
\end{figure}

\textbf{Worst cases} expose clinical risk. Baselines that perform 
well in-domain---TransUNet, BEAL, and DoFE---produce anatomically 
impossible predictions where the cup escapes the disc (Fig. \ref{fig:special}), 
rendering the vCDR uninterpretable. NPS-Net eliminates this failure by construction.

Under \textbf{extreme conditions} (severe illumination degradation),
all baselines fail.  NPS-Net recovers plausible boundaries, sustained
by the entropy gate shifting toward the global shape prior as local
appearance degrades.

% =========================================================================
\section{Discussion}
\label{sec:discussion}
% =========================================================================

\subsection{Representation as the Primary Lever}

The central finding is that the output representation---not the
architecture, loss function, or training strategy---is the primary
determinant of structural reliability under domain shift.  All seven
baselines share a common failure mode: unconstrained Cartesian masks
with soft training-time signals to encourage anatomical plausibility.
In-domain, this suffices---DoFE leads on Cup Dice and Rim Corr.
Under severe domain shift, these soft constraints collapse: DoFE
degrades below vanilla UNet on vCDR MAE (RIM-ONE), and BEAL's Cup
Dice falls to 0.6782 with $\pm$0.247 standard deviation.  PAPILA
reinforces this pattern on a distinct axis: Cartesian methods produce
Disc Dice as low as 0.36 and \emph{negative} Rim Corr.

NPS-Net avoids this by changing what the network is allowed to
predict.  The monotone construction~\eqref{eq:monotone} restricts
outputs to radially non-increasing functions, excluding fragmented
predictions by definition.  Factorized nesting~\eqref{eq:nesting}
restricts the cup to a multiplicative fraction of the disc, excluding
topology violations.  These are exact invariants that hold
for every input, parameter configuration, and domain.

\subsection{Cross-Domain vs.\ Zero-Shot Performance}

The DRISHTI-GS + REFUGE results (Table~\ref{tab:drishti_refuge})
reveal an instructive intermediate regime.  Under moderate domain
shift, Cartesian baselines retain competitive overlap scores---UNet
leads Cup Dice at 0.8603---yet their clinical geometry metrics
already degrade: BEAL's vCDR MAE rises to 0.1440, over triple its
internal-test value.  NPS-Net sweeps all three clinical metrics (vCDR
MAE 0.0636, Rim MAE 0.0531, Rim Corr 0.6856), demonstrating that
angular fidelity benefits from geometric structure even before domain
shift becomes severe.  This progressive degradation pattern confirms
that the representational advantage of polar occupancy estimation
compounds with increasing distributional distance: decisive on
clinical metrics for DRISHTI-GS + REFUGE, overwhelming on RIM-ONE and
PAPILA.

\subsection{Entropy-Gated Adaptation Without Parameters}

Unlike test-time adaptation methods requiring iterative entropy
minimization~\cite{bib24} or shape moment matching~\cite{bib25}---both
modifying parameters at inference and risking catastrophic
forgetting---our confidence gate~\eqref{eq:gate} modulates the shape
prior within a single forward pass.  The ablation confirms its
specific contribution: the largest single-step Rim Corr improvement
(0.43$\to$0.52) occurs when the shape prior activates, indicating
that angular fidelity benefits most from geometric context where
local appearance is ambiguous.

\subsection{Clinical Implications}

The most consequential property for deployment is not any single
metric improvement but the elimination of structurally invalid
outputs.  A screening system producing a cup-outside-disc prediction
yields undefined vCDR---a result that cannot be triaged, compared
longitudinally, or acted upon.  By guaranteeing
$P_c\!\leq\!P_d$, NPS-Net provides a lower bound on
output interpretability: every prediction is clinically meaningful.
The rim-profile correlation (0.56 on RIM-ONE,
0.50 on PAPILA, 0.69 on DRISHTI-GS + REFUGE, vs.\ $\leq$0.31 for
all baselines on zero-shot sets) indicates that NPS-Net preserves the
angular distribution of thinning---the spatial signature
distinguishing early focal from advanced diffuse damage.

\subsection{Limitations}

The polar representation assumes approximate knowledge of the disc
center; Polar-TTA partially relaxes this ($\pm$16\,px search), but
large localization errors require an upstream detection stage.
Polar-TTA introduces $27{\times}$ overhead; single-pass NPS-Net
maintains all guarantees at 128~FPS, and the overhead can be reduced
via coarse-to-fine search.  Evaluation on modalities not represented
in training (ultra-widefield, smartphone fundoscopy) remains open.

% =========================================================================
\section{Conclusion}
\label{sec:conclusion}
% =========================================================================

The proposed NPS-Net formulates OD/OC segmentation as nested radially monotone
polar occupancy estimation, encoding star-convexity, nested
containment, and angular rim fidelity as exact invariants of
the output parameterization rather than soft training-time penalties.
On RIM-ONE under strict zero-shot conditions, NPS-Net improves Cup
Dice by 12.8\% over the strongest baseline, reduces vCDR MAE by 56\%,
and maintains 100\% anatomical validity.  On PAPILA, Disc Dice
reaches 0.9438 with Disc HD95 of 2.78\,px.  On DRISHTI-GS + REFUGE,
NPS-Net achieves the best vCDR MAE (0.0636) and Rim Correlation
(0.6856), sweeping all clinical geometry metrics.  These results
establish a principle with implications beyond glaucoma screening:
when target structures possess known geometric invariants, encoding
them into the output representation yields robustness that
training-time optimization alone cannot achieve.

\section{Data and Code Availability}
The source code for this work is publicly available at \url{https://github.com/Rimsa78/NPS-Net.git}. The dataset used in this study is an open-source dataset. 

% =========================================================================

\end{document}